\documentclass{article}

\usepackage{booktabs, tabularx, makecell}
\usepackage{times}
\usepackage{latexsym}
\usepackage{graphicx} 
\usepackage{float} 
\usepackage{subfigure} 
\usepackage{amsmath,amssymb}
\usepackage{multirow}
\usepackage{algorithm}
\usepackage{caption}       
\usepackage{hyperref}
\usepackage{placeins}
\usepackage{url}
\usepackage{xcolor}
\usepackage{algorithmicx}

\usepackage{capt-of}
\usepackage[many]{tcolorbox}
\tcbset{cgpoalgo/.style={
  breakable, enhanced,
  colback=white, colframe=black,
  left=0.7em, right=0.7em, top=0.5em, bottom=0.5em,
  boxrule=0.4pt, arc=2pt, before skip=6pt, after skip=6pt
}}

\usepackage{microtype}
\usepackage{graphicx}
\usepackage{subcaption}

\usepackage{tabularx}
\usepackage[preprint]{icml2026}

\usepackage{booktabs}
\usepackage{multirow}
\usepackage[table]{xcolor}
\usepackage{pifont}

\usepackage{amsmath}
\usepackage{amssymb}
\usepackage{mathtools}
\usepackage{amsthm}
\newcommand{\MethodName}{CGPO}
\usepackage{algpseudocode}  
\usepackage[capitalize,noabbrev]{cleveref}

\theoremstyle{plain}

\theoremstyle{definition}

\theoremstyle{remark}

\usepackage[textsize=tiny]{todonotes}

\icmltitlerunning{Enhancing LLM Reasoning via Non-Human-Like Reasoning Path Preference Optimization}

\begin{document}

\twocolumn[
  \icmltitle{Enhancing LLM Reasoning via \\ Non-Human-Like Reasoning Path Preference Optimization}

  \icmlsetsymbol{equal}{*}

  \begin{icmlauthorlist}
    \icmlauthor{Junjie Lu}{equal,uts}
    \icmlauthor{Yuliang Liu}{equal,sii,nju}
    \icmlauthor{Chaofeng Qu}{stu}
    \icmlauthor{Wei Shen}{IR}
    \icmlauthor{Zhouhan Lin}{sii,sjtu}
    \icmlauthor{Chuheng Zhang}{msra}
    \icmlauthor{Min Xu}{uts}

  \end{icmlauthorlist}

  \icmlaffiliation{uts}{University of Technology Sydney}
  \icmlaffiliation{sii}{Shanghai Innovation Institute}
  \icmlaffiliation{stu}{Southeast University}
  \icmlaffiliation{msra}{Microsoft Research}
  \icmlaffiliation{nju}{Nanjing University}
  \icmlaffiliation{sjtu}{Shanghai Jiao Tong University}
  \icmlaffiliation{IR}{Independent Researcher}

  \icmlcorrespondingauthor{Junjie Lu}{Junjie.Lu@uts.edu.au}
  \icmlcorrespondingauthor{Min Xu}{Min.Xu@uts.edu.au}

  \vskip 0.3in
]

\printAffiliationsAndNotice{\icmlEqualContribution}

\begin{abstract}
    Current approaches for strengthening LLM reasoning tend to introduce a training bias toward human-like reasoning trajectories. In step-wise preference optimization, in particular, dependence on human or higher-capacity model annotations for intermediate steps limits exploration of alternative, non-human-like reasoning paths and thus constrains achievable performance. Furthermore, through a small-scale pilot study, we observed that in approximately 75\% of cases, the model's first erroneous step occurs after the lowest-confidence point. This suggests that guiding the model at its lowest-confidence point before an error provides more accurate supervision than locating the first explicit error. In this paper, we propose \textbf{C}onfidence-\textbf{G}uided Reasoning Path \textbf{P}reference \textbf{O}ptimization (CGPO), a method that leverages a confidence signal to identify points of maximal uncertainty in the model’s reasoning process and applies self-generated, non-human-like reasoning-path guidance to mitigate trajectory drift. Our experiments span diverse models applied to both code and mathematical reasoning tasks. The results show that, with the same amount of training data, our method using data generated by a small model can achieve better performance in most cases compared with approaches using data generated by a strong model or human-annotated.
\end{abstract}

\section{Introduction}

Large language models (LLMs) have demonstrated remarkable capabilities across various domains, stemming from their ability to reason~\citep{hao2023reasoning,lewkowycz2022solving,magister2022teaching,khalifa2025processrewardmodelsthink}.
To further enhance their reasoning capabilities, researchers have conducted extensive studies on both the training and inference phases of the models. Training methods typically focus on teaching the model to follow human instruction through supervised training~\citep{yue2023mammoth,luo2023wizardcoder}, or on preference optimization of reasoning chains using reward models or human feedback, usually known as RLHF~\citep{Guo2025DeepSeekR1,zhang2024rest}. 
For inference methods, researchers often employ prompt tuning~\citep{kojima2022large,imani2023mathprompter}, or intervene in the model’s generation process by assessing consistency~\citep{wang2022self,chen2022program,taubenfeld2025confidence}, predicting confidence~\citep{weng2022large,zhu2025uncertainty}, and leveraging expected outcomes~\citep{madaan2023self,cheng2024spar}.

While chain-of-thought (CoT)~\citep{wei2023chainofthoughtpromptingelicitsreasoning} has become the standard approach for addressing complex reasoning tasks, early training methods primarily focused on outcome optimization, with limited attention to the structure and quality of reasoning paths~\citep{illusion-of-thinking}. One possible reason is that outcome-based approaches have yielded reasonably satisfactory performance on current benchmarks~\citep{gao2024designing,hosseini2024v}, while another is that assigning fine-grained rewards remains costly. Nevertheless, there have been efforts to optimize finer-grained aspects of the reasoning process, like step-wise methods~\citep{snell2024scaling,wang2024multi} and token-wise methods~\citep{zhang2024generative,lin2024critical}.

\begin{figure*}[t] 
\centering 
\includegraphics[width=1.0\textwidth,height=0.36\textheight]{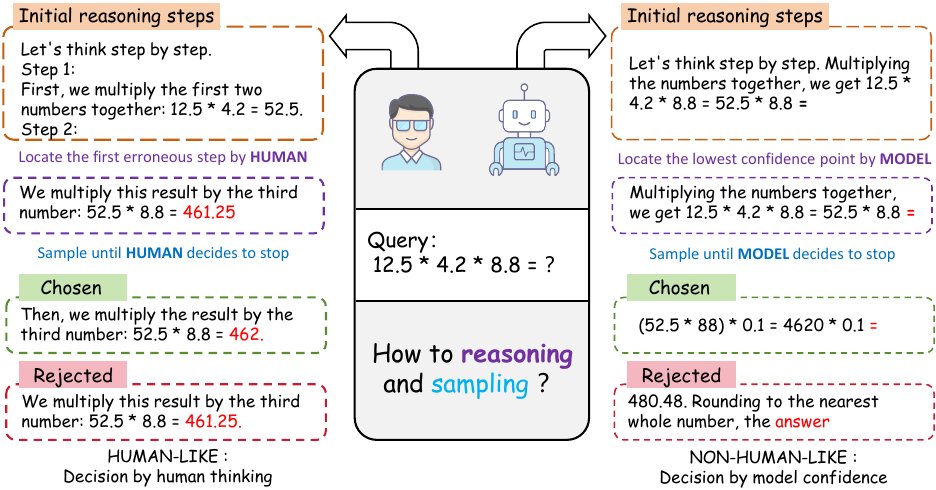} 
\caption{The difference between human-like reasoning and non-human-like reasoning. Human-like methods emulate human cognitive processes and typically use predefined, readable anchors as delimiters to split the reasoning path. Nevertheless, it imposes substantial constraints on the content and format of data during sampling. In contrast, in this paper, non-human-like delineate reasoning steps are based on model confidence and provide reward supervision for tokens where the model is most uncertain, incurring lower cost and exhibiting broader applicability across various tasks.}
\label{fig:humanlike_nonhumanlike} 
\end{figure*}

Existing step-wise preference optimization approaches usually locate the model's first error~\citep{luo2024improve,lu2024step} and construct preference pairs accordingly, enabling focused and targeted optimization. 
However, these methods often involve substantial labeling overhead, posing a bottleneck in their practical applicability. 
Furthermore, our analysis on the incorrect answers of multiple models on mathematical reasoning tasks reveals that models often exhibit confusion, indicated by relatively low confidence in generation, well before about 75\% of the actual errors. 
Examples are provided in Appendix~\ref{app:case_conf_error}. 
This observation inspired us to develop a method capable of identifying issues earlier in a fully automated, human-free manner.

We propose Confidence-Guided Reasoning Path Preference Optimization (CGPO), which anchors preference optimization to model uncertainty rather than human alignment or process-error identification.
Concretely, CGPO is grounded in two key principles:
(1) enabling the model to identify and refine its own confusion, and
(2) leveraging preference pair construction to guide optimization. By preference optimization, our approach introduces a thoroughly non-human-like reasoning path exploration, as shown in Figure~\ref{fig:humanlike_nonhumanlike}.

Our experiments show that CGPO yields consistent gains on mathematical reasoning under the same number of training samples. Specifically, averaged across all evaluated models, CGPO yields a performance gain of 2.1\% on GSM8K~\citep{cobbe2021trainingverifierssolvemath} and 1.2\% on MATH~\citep{hendrycksmath2021} relative to the baseline. Importantly, our method uses the training data generated by the policy model itself, whereas the baseline uses the training data generated by stronger models or annotated by humans. Models trained with CGPO further demonstrate better generalization on CNMO24\citep{liu2024your}. For the code generation task, we train the DeepSeek-Coder-Instruct-7B by CGPO and observe a 0.7\% improvement on LiveCodeBench~\citep{jain2024livecodebenchholisticcontaminationfree} and 0.9\% improvement on LeetCodeDataset~\citep{xia2025leetcodedatasettemporaldatasetrobust}. We also examine the scalability of our method by increasing the training data from 10k to 80k, which results in the data leading to continued improvements. 

Our ablation studies and analyses further investigate the differences in effects between human-like and non-human-like splits. Moreover, our experiments reveal that, within a certain range, intermediate reasoning path-focused optimization leads to more substantial enhancement than optimizing the entire reasoning path.

Our main contributions are as follows:

\begin{enumerate} 
\item We introduce CGPO, a method that automatically constructs preference data and performs reasoning path preference optimization for LLMs without relying on more powerful models or human supervision. CGPO can sample key non-human-like reasoning paths from the model to construct preference-learning data, therefore improving the model’s reasoning ability.
\item We demonstrate on several mathematical reasoning benchmarks that CGPO improves models’ reasoning ability. We also apply CGPO to code generation, showing that CGPO generalizes to broader domains where human-like reasoning paths are hard to define.
\item We build a simple and efficient training codebase for reproducing CGPO, together with multiple datasets of non-human-like reasoning paths across different models and the corresponding models trained on these data. We will release all code, datasets, and trained models to facilitate reproducibility and comparison upon acceptance of this paper.
\end{enumerate}

\section{Related Works}

\textbf{Preference Optimization for LLM:} Preference optimization methods have been investigated to improve LLM performance on complex tasks. PPO~\citep{schulman2017proximalpolicyoptimizationalgorithms} balances reward maximization with stability through a clipped objective, serving as a foundational method in RLHF but with considerable complexity, such as involving the critic training and clipping hyperparameter tuning. Subsequently, critic-free RL methods~\citep{shao2024deepseekmathpushinglimitsmathematical, ahmadian2024basicsrevisitingreinforcestyle, hu2025reinforceefficientrlhfalgorithm} reduce such complexity by using reward model outputs as immediate rewards for policy gradient updates. 
The methods like DPO~\citep{rafailov2024directpreferenceoptimizationlanguage} and SimPO~\citep{meng2024simposimplepreferenceoptimization} directly optimize the model on pairwise preference signals. The reduced computational overhead, combined with sufficiently strong performance, enables preference-based methods to achieve widespread success across a variety of domains like instruction following~\citep{wu2024thinkingllmsgeneralinstruction, cheng2025sparselfplaytreesearchrefinement}, mathematical reasoning~\citep{pang2024iterativereasoningpreferenceoptimization}, and code generation~\citep{zhang2025codedpoaligningcodemodels, zhu2025flowrlmatchingrewarddistributions}.

\textbf{Confidence-Aware Methods:} Model confidence, which is commonly used to quantify model uncertainty, is widely used in LLMs~\citep{fu2025deepthinkconfidence, Taubenfeld_2025, kang2025scalablebestofnselectionlarge,li2025confidenceneedfewshotrl,singhi2025solveverifycomputeoptimalproblem,xu2025quantifyingfairnessllmstokens}. Confidence can be interpreted from multiple perspectives. 
While some approaches measure confidence directly using the prediction probability of selected tokens~\citep{Farquhar2024-qf, liu2025adaptivestepautomaticallydividingreasoning},
others rely on the model’s outputs. 
For example, \citet{xiong2024llmsexpressuncertaintyempirical} proposed a systematic framework to express model confidence by generation, and \citet{yona2024largelanguagemodelsfaithfully} measures model confidence by consistency of answers. There are also several methods that use the entropy of probability distribution or probe-based methods to measure model confidence in other tasks~\citep{stolfo2024confidenceregulationneuronslanguage, beigi2024internalinspectori2robustconfidence}. 
In this work, we adopt a token probability-based confidence for its simplicity, generality, and efficiency.

\textbf{Reasoning Trajectory Optimization:} 
LLM reasoning is viewed as a foundational capability for improving performance on complex tasks, and numerous studies have investigated how to improve this ability~\citep{openai2024openaio1card, qwen3}.
Supervised fine-tuning methods directly align model trajectory to annotated sequence~\citep{yu2025long}; RLHF methods typically force the model to optimize the reasoning trajectory towards a higher score~\citep{yu2025dapoopensourcellmreinforcement}; and test time methods intervene in the generation process to optimize reasoning trajectories~\citep{muennighoff2025s1simpletesttimescaling}. The aforementioned training approaches, especially RLHF, are generally concerned with outcome optimization and do not explicitly enforce correctness along the reasoning trajectories. 
In contrast, \citet{lai2024stepdpostepwisepreferenceoptimization} directly optimizes an independent reasoning path, the nearest one of our method, while with a significant annotation challenge. Our approach avoids reliance on human-annotated data and instead aligns with the model’s distribution over reasoning trajectories, exhibiting stronger generalization and scalability.

\section{Methods}

\begin{figure*}[ht] 
\centering 
\includegraphics[width=1.0\textwidth, height=0.26\textheight]{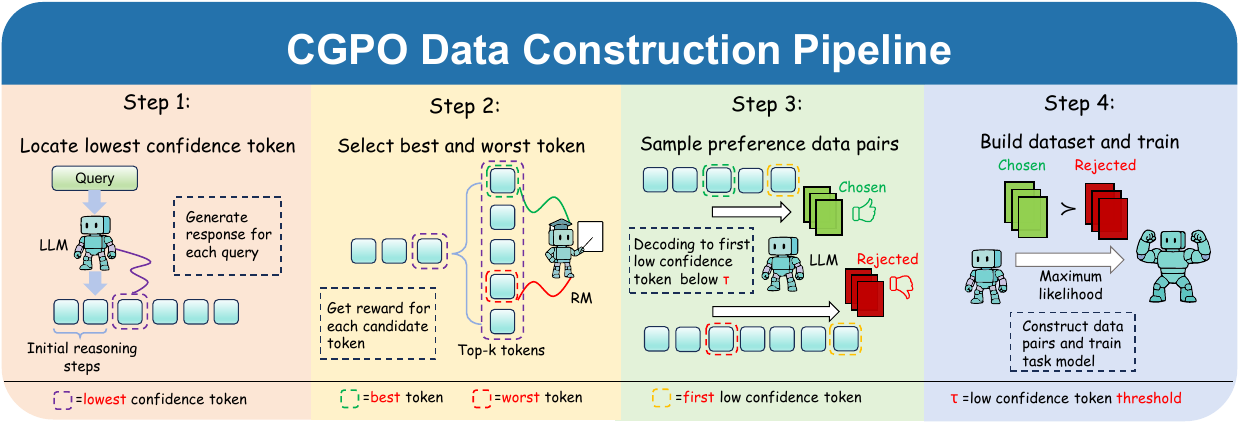} 
\caption{\MethodName \vspace{1pt} overview. The core idea of our method consists of two parts: 1) explore the non-human-like reasoning path by determining the model reasoning step using the initial policy model $\pi_\theta$, and 2) construct chosen/rejected pairs to train $\pi_\theta$ with step-wise DPO training objective.
}
\label{fig:training_data_construction_pipeline} 
\end{figure*}

In this section, we will introduce the basic formulation, the construction of the training data, and the training objective of \MethodName.
We present the overall pipeline in Figure~\ref{fig:training_data_construction_pipeline}.

\subsection{Preliminaries}

Given a problem $\mathbf{x}$, the initial policy model $\pi_\theta$ generates a reasoning sequence
$\mathbf{y} = (y_{1},y_{2},\dots,y_{T}),\, y_{t}\in\mathcal{V}$, where \(\mathcal{V}\) denotes the vocabulary.
Notably, the selected token in generation may not be the token with the top-1 probability since we use a temperature-based sampling strategy for exploration. Let $z_t(v)$ denote the logit of token $v$ at step $t$. 
The sampling distribution with temperature $\beta_{\text{smpl}}$ is defined as:
\begin{equation}
p_{\beta_{\text{smpl}}}(v \mid \pi_\theta, \mathbf{x}, y_{<t})
= \frac{\exp(z_t(v) / \beta_{\text{smpl}})}
{\sum_{v' \in \mathcal{V}} \exp(z_t(v') / \beta_{\text{smpl}})}.
\end{equation}
In addition, we define the model confidence $c_t$ as:

\begin{equation}
c_t \triangleq p_{\beta_{\text{smpl}}}(y_t \mid \pi_\theta, \mathbf{x}, y_{<t}).
\end{equation}

\subsection{Preference Pairs Construction}

\textbf{Step Definition.} 
To construct non-human-like step-wise chosen/rejected preference pairs, we first need to define non-human-like steps that differ from those used in previous step-wise optimization methods.

Given $\mathbf{x}$ and the corresponding response $\mathbf{y}$, we can get the confidence sequence for $\mathbf{y}$, denoted by $\mathbf{c}=(c_1, c_2, \dots, c_T)$. A confidence threshold is employed to split the reasoning steps, with $\tau$ determined based on the distribution of all confidence values $c$ across the dataset. 
We define split points as token positions where confidence falls below the threshold $\tau$, $\mathcal{T} = \{ t \mid c_t < \tau \}$.
Let $\{t_j\}_{j=1}^J$ be the sorted elements of $\mathcal{T}$, with $t_0 = 1$ and
$t_{J+1} = T+1$. Each reasoning step is defined as a contiguous token subsequence:
\begin{equation}
\mathbf{s}_j := (y_{t_{j-1}}, \dots, y_{t_j - 1}), \quad j = 1,\dots,J+1.
\end{equation}
Thus, the generated sequence is segmented into steps
$\mathbf{s} = (s_1, \dots, s_{J+1})$.

\textbf{Initial Reasoning Trajectory Determination.}
To align with the pair-wise preference optimization paradigm, we select a shared
initial reasoning trajectory before an uncertain decision point. 
Let
\begin{equation}
t^* = \arg\min_{t \in \{1,\dots,T\}} c_t
\end{equation}
be the index of the least-confident token. We define the shared initial reasoning
trajectory as the prefix before $t^*$:
\begin{equation}
\mathbf{s}_{\text{init}} := (y_1,\dots,y_{t^*-1}).
\end{equation}

\textbf{Chosen/Rejected Pairs Construction.} We introduce a reward model $R$ to select the first token of the chosen/rejected pairs given $(\mathbf{x}, \mathbf{s}_{init})$. The top-k candidate tokens, ordered by their logits, are denoted as $(v_1, v_2, \dots,v_k)$. The reward for each token is obtained as $R(\mathbf{x}, \mathbf{s}_{init}, v_i)$. Then, we select the token with the highest score and the one with the lowest score as the starting tokens of the chosen/rejected pairs,
denoted by $y^+$ and $y^-$ respectively.

Subsequently, $\pi_\theta$ samples continuation sequences conditioned on
$(\mathbf{x}, \mathbf{s}_{\text{init}}, y^+)$ and $(\mathbf{x}, \mathbf{s}_{\text{init}}, y^-)$
until the end-of-sequence token or the first token with confidence below $\tau$.
Formally, we define
\begin{align}
\mathbf{s}^+ &\sim \pi_\theta(\cdot \mid \mathbf{x}, \mathbf{s}_{\text{init}}, y^+), \\
\mathbf{s}^- &\sim \pi_\theta(\cdot \mid \mathbf{x}, \mathbf{s}_{\text{init}}, y^-),
\end{align}
as the chosen and rejected continuation sequences, respectively.
Finally, for each given input, we have two reasoning trajectories: the chosen one $(\mathbf{s}_{\text{init}}, \mathbf{s}^+)$ and the rejected one $(\mathbf{s}_{\text{init}}, \mathbf{s}^-)$.

\subsection{Training Objective}

Our goal is to optimize $\pi_\theta$ based on the triplets $(\mathbf{s}_{init}, \mathbf{s}^+, \mathbf{s}^-)$. Let \(\mathcal{D}\) denote the collection of such triples. With a fixed reference model \(\pi_{\text{ref}}\) and temperature \(\beta>0\), we optimize $\pi_\theta$ using the DPO-like objective

\begin{align}
\mathcal{L}_{\text{CGPO}}(\theta) 
&= -\,\mathbb{E}_{(\mathbf{s}_{init}, \mathbf{s}^+, \mathbf{s}^-)\sim\mathcal{D}} 
\Big[ \log \sigma\big( \beta(\Delta) \big) \Big], \\
\sigma(a) &= \frac{1}{1+e^{-a}}.
\end{align}

where \(\beta\) is the inverse temperature and

\begin{align}
\Delta &= \Delta_\theta - \Delta_{\text{ref}}, \\
\Delta_\theta &\triangleq \log \pi_{\theta}(\mathbf{s}^+\!\mid \mathbf{s}_{init}) 
                 - \log \pi_{\theta}(\mathbf{s}^-\!\mid \mathbf{s}_{init}), \\
\Delta_{\text{ref}} &\triangleq \log \pi_{\text{ref}}(\mathbf{s}^+\!\mid \mathbf{s}_{init}) 
                     - \log \pi_{\text{ref}}(\mathbf{s}^-\!\mid \mathbf{s}_{init}).
\end{align}

Sequence log-likelihoods use natural logs and decompose autoregressively as
\begin{align}
\log \pi(s^{\pm}\!\mid \mathbf{s}_{init})
\ =\
\sum_{t}^{seq}
\log \pi\!\big(y_{t}^{\pm}\mid x,\mathbf{s}_{init}).
\end{align}

where the sum starts at the branching token $y^\pm$.
This objective increases the likelihood of the high-quality trajectory $\mathbf{s}^+$ while decreasing that of $\mathbf{s}^-$, which are selected by the previously defined reward model $R$; the reference term \(\Delta_{\text{ref}}\) stabilizes the relative preference signal. 
We use Algorithm~\ref{alg:cgpo} to demonstrate the entire pipeline of CGPO.

\begin{algorithm}[t]
\caption{\MethodName\vspace{1pt} Pipeline}\label{alg:cgpo}
\begin{algorithmic}[1]
\Require Dataset $\mathcal{X}$, $\pi_\theta$, $R$, temp, $\tau$, $k$, $\beta$
\Ensure Trained policy $\pi_\theta$
\State Initialize $\pi_\theta$, preference dataset $\mathcal{D}\gets \emptyset$
\For{each input $\mathbf{x}\in\mathcal{X}$}
  \State Sample $\mathbf{y}=(y_1, \cdots, y_T)$ from $\pi_\theta$
  \State Record confidence $\{c_t=\pi_\theta(y_t\mid \mathbf{x}, \mathbf{y}_{<t})\}_{t=1}^T$
  \State Split $\mathbf{y}$ into steps $\textbf{s}=(\textbf{s}_1,\dots,\textbf{s}_J)$ using $\tau$
  \State Find the most uncertain step $t^*:=arg\min_t c_t$
  \State Set $\textbf{s}_{\text{init}}=(y_1,\dots,y_{t^*})$
  \State Obtain top-$k$ next-token candidates $(v_1,\dots,v_k)$
  \State Select $i^{+/-} = arg \max/\min_i R(\mathbf{x}, s_{\text{init}}, v_i)$

  \State Choose $y^+ = v_{i^+}$ and $y^- = v_{i^-}$

  \State Sample $\textbf{s}^+$, $\textbf{s}^-$ using $\pi_\theta$ starting at $y^+$, $y^-$, stopping at $\langle$EOS$\rangle$ or the first token with confidence $<\tau$

  \State Add $(\mathbf{x}, \textbf{s}_{\text{init}}, \textbf{s}^+, \textbf{s}^-)$ to $\mathcal{D}$

\EndFor

\For{each $(\mathbf{s}_{\text{init}}, \mathbf{s}^+, \mathbf{s}^-)\in \mathcal{D}$}
  \State Update $\pi_{\theta}$ using the CGPO objective in Eq.~(9)
\EndFor
\State \Return $\pi_\theta$
\end{algorithmic}
\end{algorithm}
\vspace{-0.5em}

\section{Experiments and Analysis}

In this section, we first introduce our experimental setup, including datasets, models, baselines, metrics, and parameter settings. We then present the experimental results, followed by an analysis of the transferability, generalization, and scalability of CGPO.

\subsection{Experiments Setup}

\textbf{Datasets and Models:} 
For mathematical reasoning tasks, we use the prompts from the Step-DPO-10k\footnote{https://huggingface.co/datasets/xinlai/Math-Step-DPO-10K} dataset as the model inputs $\mathbf{x}$ to construct the preference dataset for training. 
We evaluate CGPO with various models serving as the policy model $\pi_\theta$, including MetaMath-Mistral-7B \citep{yu2024metamathbootstrapmathematicalquestions}, MetaMath-Llama-8B \citep{liu2025adaptivestepautomaticallydividingreasoning}, Qwen2-7B-SFT, Qwen1.5-32B-SFT, DeepSeekMath-Base-SFT\cite{lai2024stepdpostepwisepreferenceoptimization}, DeepSeek-R1-Distill-Qwen-1.5B, and DeepSeek-R1-Distill-Qwen-7B\cite{Guo2025DeepSeekR1}.
To ensure that the capabilities of the reward model (RM) $R$ align with the policy model to provide accurate token-level reward, to prevent performance gains from an overly strong RM, We adopt ASPRM-L\footnote{https://huggingface.co/Lux0926/ASPRM-L}, whose base model is competitive with our policy model while enabling fine-grained evaluation, as the RM to select candidate tokens after $\textbf{s}_{init}$. For evaluation, we benchmark standard instruction models on the GSM8K and MATH test sets, while reasoning models are evaluated on GSM8K and MATH500 to enhance evaluation efficiency.

For code generation tasks, we use the prompts from the training set of the LeetCodeDataset\footnote{https://huggingface.co/datasets/newfacade/LeetCodeDataset} as model inputs $\mathbf{x}$ to construct the preference dataset. The policy model $\pi_\theta$ is Deepseek-Coder-7B-Instruct-v1.5 \citep{guo2024deepseekcoderlargelanguagemodel}, and we adopt ASPRM-D\footnote{https://huggingface.co/Lux0926/ASPRM-D} as our reward model to select candidate tokens. The trained models are evaluated on the test sets of LiveCodeBench-v6\footnote{https://huggingface.co/livecodebench} and LeetCodeDataset.

\textbf{Baselines and Metrics:} For the mathematics domain, five training settings are compared: the base model, Step-DPO, CGPO, DPO trained on the sampled full chosen and rejected sequence pairs constructed by Step-DPO (Step-DPO data), and DPO trained on the sampled full chosen and rejected sequence pairs constructed by CGPO (CGPO data). For the code domain, three training settings are compared: the base model, DPO trained on CGPO data, and CGPO.

For all tasks, large reasoning models follow prior work\cite{openr1} by sampling 4 generations with temperature $0.6$ and top-$p$ $0.95$, and reporting pass@1 for comparison; all other models use greedy decoding and are compared by directly computed accuracy. In mathematical reasoning tasks, a prediction is considered correct only if model’s final answer exactly matches the ground truth. For code-generation tasks, the generated code is executed in a sandbox and deemed correct if it passes all test cases.

\textbf{Parameter Settings:} For each base model in the mathematical domain, we sample one time per prompt from the dataset, yielding 10,795 outputs. For each output, we locate the lowest confidence point and use the reward model to score the top 8 token candidates, then branch into two reasoning paths. 
The policy models decode tokens until the confidence falls below 2\% of its confidence distribution or reach EOS token. 
This produces a preference dataset of 10,795 pairs for each policy model, equivalent in size to Step-DPO-10k. 
For the code domain, we sample 4 times per prompt from the 2,641 LeetCodeDataset training data and apply the same token selection and reward-model procedure to generate 10,564 preference pairs. We follow the Step-DPO training configuration to reproduce baselines. Hyperparameter settings for CGPO are provided in Appendix~\ref{sec:training_details}.

\begin{table*}[t]
\centering
\caption{Performance comparison of various models on mathematical and code benchmarks. \textcolor{red!50!white}{Red} and \textcolor{green!50!black}{Green} numbers represent the performance improvement or decline compared to the base model.}
\renewcommand{\arraystretch}{1.25}
\vspace{6pt}

\begin{tabular}{lcccccc}
\toprule
\multirow{2}{*}{\textbf{Dataset}} 
& \multirow{2}{*}{\textbf{Model}}  
& \multirow{2}{*}{\textbf{Base}}  
& \multicolumn{2}{c}{\textbf{+ DPO}} 
& \multirow{2}{*}{\textbf{+ Step-DPO}} 
& \multirow{2}{*}{\textbf{+ CGPO}} \\
\cmidrule(lr){4-5}
& & & \scriptsize Step-DPO DATA & \scriptsize CGPO DATA & & \\

\midrule
\multirow{5}{*}{\textbf{GSM8K}}

& MetaMath-Mistral-7B 
& 77.1 
& 76.5 \textcolor{green!50!black}{ (-0.6)}
& 78.5 \textcolor{red!50!white}{ (+1.4)}
& 76.7 \textcolor{green!50!black}{ (-0.4)}
& \textbf{81.1} \textcolor{red!50!white}{ (+4.0)} \\ 

& MetaMath-Llama-8B 
& 81.8 
& 85.0 \textcolor{red!50!white}{ (+3.2)}
& 86.0 \textcolor{red!50!white}{ (+4.2)}
& 84.8 \textcolor{red!50!white}{ (+3.0)}
& \textbf{86.4} \textcolor{red!50!white}{ (+4.6)} \\

& Qwen2-7B-SFT 
& 87.9 
& 85.0 \textcolor{green!50!black}{ (-2.9)}
& 88.8 \textcolor{red!50!white}{ (+0.9)}
& 88.4 \textcolor{red!50!white}{ (+0.5)}
& \textbf{89.8} \textcolor{red!50!white}{ (+1.9)} \\

& Qwen1.5-32B-SFT 
& 90.0 
& 90.3 \textcolor{red!50!white}{ (+0.3)}
& 90.1 \textcolor{red!50!white}{ (+0.1)}
& 90.3 \textcolor{red!50!white}{ (+0.3)}
& \textbf{90.6} \textcolor{red!50!white}{ (+0.6)} \\

& DeepSeekMath-Base-7B-SFT 
& 86.3 
& 87.0 \textcolor{red!50!white}{ (+0.7)}
& 86.9 \textcolor{red!50!white}{ (+0.6)}
& 86.5 \textcolor{red!50!white}{ (+0.2)}
& \textbf{87.4} \textcolor{red!50!white}{ (+1.1)} \\

& DeepSeek-R1-Distill-Qwen-1.5B 
& 76.7
& 77.2 \textcolor{red!50!white}{ (+0.5)}
& 76.3 \textcolor{green!50!black}{ (-0.4)}
& 78.6  \textcolor{red!50!white}{ (+1.9)}
& \textbf{79.0} \textcolor{red!50!white}{ (+2.3)} \\

& DeepSeek-R1-Distill-Qwen-7B
& 86.1 
& \textbf{86.7} \textcolor{red!50!white}{ (+0.6)}
& 85.6 \textcolor{green!50!black}{ (-0.5)}
& 86.2 \textcolor{red!50!white}{ (+0.1)}
& 86.5 \textcolor{red!50!white}{ (+0.4)} \\

& Average
& 83.7 
& 84.0 \textcolor{red!50!white}{ (+0.3)}
& 84.6 \textcolor{red!50!white}{ (+0.9)}
& 84.5 \textcolor{red!50!white}{ (+0.8)}
& \textbf{85.8} \textcolor{red!50!white}{ (+2.1)} \\

\midrule

\multirow{5}{*}{\textbf{MATH}}

& MetaMath-Mistral-7B 
& 26.6 
& 26.9 \textcolor{red!50!white}{ (+0.3)}
& 27.0 \textcolor{red!50!white}{ (+0.4)}
& 27.3 \textcolor{red!50!white}{ (+0.7)}
& \textbf{27.4} \textcolor{red!50!white}{ (+0.8)} \\ 

& MetaMath-Llama-8B 
& 39.3 
& 40.1 \textcolor{red!50!white}{ (+0.8)}
& 40.6 \textcolor{red!50!white}{ (+1.3)}
& 40.2 \textcolor{red!50!white}{ (+0.7)}
& \textbf{40.7} \textcolor{red!50!white}{ (+1.4)} \\

& Qwen2-7B-SFT 
& 54.2 
& 51.7 \textcolor{green!50!black}{ (-2.5)}
& 54.5 \textcolor{red!50!white}{ (+0.3)}
& 54.6 \textcolor{red!50!white}{ (+0.4)}
& \textbf{55.0} \textcolor{red!50!white}{ (+0.8)} \\

& Qwen1.5-32B-SFT 
& 54.2 
& 54.4 \textcolor{red!50!white}{ (+0.2)}
& 55.3 \textcolor{red!50!white}{ (+1.1)}
& 55.5 \textcolor{red!50!white}{ (+1.3)}
& \textbf{55.7} \textcolor{red!50!white}{ (+1.5)} \\

& DeepSeekMath-Base-7B-SFT 
& 51.4 
& 51.8 \textcolor{red!50!white}{ (+0.4)}
& 51.9 \textcolor{red!50!white}{ (+0.5)}
& 52.5 \textcolor{red!50!white}{ (+1.1)}
& \textbf{52.8} \textcolor{red!50!white}{ (+1.4)} \\ 

& DeepSeek-R1-Distill-Qwen-1.5B 
& 82.2
& 82.5 \textcolor{red!50!white}{ (+0.3)}
& 82.3 \textcolor{red!50!white}{ (+0.1)}
& 81.7 \textcolor{green!50!black}{ (-0.5)}
& \textbf{83.7} \textcolor{red!50!white}{ (+1.5)} \\

& DeepSeek-R1-Distill-Qwen-7B
& 93.2 
& 93.5 \textcolor{red!50!white}{ (+0.3)}
& 93.3 \textcolor{red!50!white}{ (+0.1)}
& 93.3 \textcolor{red!50!white}{ (+0.1)}
& \textbf{94.0} \textcolor{red!50!white}{ (+0.8)} \\
                    
& Average
& 57.3
& 57.3 \textcolor{red!50!white}{ (+0.0)}
& 57.8 \textcolor{red!50!white}{ (+0.5)}
& 57.9 \textcolor{red!50!white}{ (+0.6)}
& \textbf{58.5} \textcolor{red!50!white}{ (+1.2)} \\                  
\midrule

\textbf{LiveCodeBench} 
& Deepseek-Coder-7B-Instruct-v1.5 
& 19.3 
& / 
& 19.6 \textcolor{red!50!white}{ (+0.3)}
& / 
& \textbf{20.0} \textcolor{red!50!white}{ (+0.7)} \\

\midrule

\textbf{LeetCodeDataset} 
& Deepseek-Coder-7B-Instruct-v1.5 
& 12.7 
& / 
& 12.8 \textcolor{red!50!white}{ (+0.1)}
& / 
& \textbf{13.6} \textcolor{red!50!white}{ (+0.9)} \\

\bottomrule

\end{tabular}
\label{tab:overall_model_performance}
\end{table*}

\subsection{Overall Results}

We show the main results in Table~\ref{tab:overall_model_performance} and make the following observations:
1) For the mathematics reasoning task, CGPO achieves the highest average performance improvement across all models. Specifically, on GSM8K, CGPO yields the strongest performance for all evaluated base models except DeepSeek-R1-Distill-Qwen-7B, while on MATH, CGPO consistently achieves the best performance across all models. Compared to the base model, CGPO improves the average accuracy by 2.1\% on GSM8K and 1.2\% on MATH, indicating stable gains across different model families. Notably, Step-DPO leads to performance drops when applied to weaker models such as MetaMath-Mistral-7B and DeepSeek-R1-Distill-Qwen-1.5B, suggesting that alignment to human-like reasoning paths imposes a higher capability requirement and may yield suboptimal preference signals when the model struggles to follow such trajectories. 2) In addition, directly training DPO on full trajectories sampled by Step-DPO or CGPO can also degrade performance for certain models, possibly because sequence-level preference learning diffuses the supervision signal over the entire trajectory and fails to emphasize the critical reasoning bottlenecks. In contrast, CGPO produces consistent improvements over the base model across all evaluated settings, supporting the hypothesis that allowing models to explore and learn from non-human-like reasoning paths provides a more effective and optimizable search space than human-like reasoning paths. 3) Moreover, DPO trained on CGPO data generally outperforms its counterpart trained on Step-DPO data and, in several cases, approaches the gains of Step-DPO, further indicating that non-human-like, confidence-guided branching yields higher-quality preference data than alignment to human-like reasoning paths.

Additionally, for the code generation task, DeepSeek-Coder-7B-Instruct-v1.5 trained with CGPO achieved 0.7\% improvement on LiveCodeBench and 0.9\% improvement on the LeetCodeDataset. This demonstrates that CGPO can enhance model capabilities beyond mathematical reasoning tasks, including code generation. In such tasks, where human-like reasoning paths are not easily defined, CGPO enables the model to explore non-human-like reasoning paths and improves performance.

\FloatBarrier
\begin{table*}[t]
  \centering
  \caption{Performance comparison on out-of-domain dataset (CNMO24).}
  \renewcommand{\arraystretch}{1.25}
  \vspace{6pt}

  \begin{tabular}{l l c c c c c}
    \toprule
    \multirow{2}{*}{\textbf{Dataset}} 
    & \multirow{2}{*}{\textbf{Model}}  
    & \multirow{2}{*}{\textbf{Base}}  
    & \multicolumn{2}{c}{\textbf{+ DPO}} 
    & \multirow{2}{*}{\textbf{+ Step-DPO}} 
    & \multirow{2}{*}{\textbf{+ CGPO}} \\
    \cmidrule(lr){4-5}
    & & & \scriptsize Step-DPO DATA & \scriptsize CGPO DATA & & \\
    \midrule

    \multirow{5}{*}{\textbf{CNMO24}} 
    & MetaMath-Mistral-7B      
      & 0.0 
      & 0.0  \textcolor{red!50!white}{(+0.0)}
      & 0.0  \textcolor{red!50!white}{(+0.0)}
      & 0.0  \textcolor{red!50!white}{(+0.0)}
      & \textbf{2.8} \textcolor{red!50!white}{(+2.8)} \\

    & MetaMath-Llama-8B        
      & 0.0 
      & 0.0  \textcolor{red!50!white}{(+0.0)}
      & 0.0  \textcolor{red!50!white}{(+0.0)}
      & 0.0  \textcolor{red!50!white}{(+0.0)}
      & \textbf{2.8} \textcolor{red!50!white}{(+2.8)} \\

    & Qwen2-7B-SFT             
      & 5.6 
      & 0 \textcolor{green!50!black}{ (-5.6)}
      & 5.6  \textcolor{red!50!white}{(+0.0)}
      & 5.6  \textcolor{red!50!white}{(+0.0)}
      & \textbf{11.1} \textcolor{red!50!white}{(+5.5)} \\

    & DeepSeekMath-Base-7B-SFT 
      & 5.6 
      & 0  \textcolor{green!50!black}{ (-5.6)}
      & 5.6 \textcolor{red!50!white}{(+0.0)}
      & 5.6 \textcolor{red!50!white}{(+0.0)}
      & \textbf{5.6} \textcolor{red!50!white}{(+0.0)} \\

    & Qwen1.5-32B-SFT          
      & 5.6 
      & 2.8  \textcolor{green!50!black}{ (-2.8)}
      & 5.6  \textcolor{red!50!white}{(+0.0)}
      & \textbf{8.3}  \textcolor{red!50!white}{+(2.7)}
      & \textbf{8.3} \textcolor{red!50!white}{+(2.7)} \\

    & DeepSeek-R1-Distill-Qwen-1.5B 
    & 8.3
    & 8.3 \textcolor{red!50!white}{(+0.0)}
    & 8.3 \textcolor{red!50!white}{(+0.0)}
    & 5.6 \textcolor{green!50!black}{ (-2.7)}
    & \textbf{11.1} \textcolor{red!50!white}{ (+2.8)} \\
    
    & DeepSeek-R1-Distill-Qwen-7B
    & 8.3
    & 8.3 \textcolor{red!50!white}{(+0.0)}
    & 8.3 \textcolor{red!50!white}{(+0.0)}
    & 5.6 \textcolor{green!50!black}{ (-2.7)}
    & \textbf{11.1} \textcolor{red!50!white}{ (+2.8)} \\

    & Average
    & 4.8
    & 2.8 \textcolor{green!50!black}{ (-2.0)}
    & 4.8 \textcolor{red!50!white}{(+0.0)}
    & 4.4 \textcolor{green!50!black}{ (-0.4)}
    & \textbf{7.5} \textcolor{red!50!white}{(+2.8)} \\

    \bottomrule
  \end{tabular}

  \label{tab:model_performance_cnmo24}
\end{table*}

\subsection{Analysis of Out-of-Distribution Datasets}

To analyze the out-of-distribution generalization performance of CGPO and its performance on challenging multilingual mathematical tasks, the CNMO24~\citep{liu2024your} dataset is used to evaluate the models. CNMO24 is a competition-level benchmark consisting of 18 problems from the China National Mathematical Olympiad (CNMO) 2024. We evaluate models on both the Chinese and English versions of each problem, yielding 36 evaluation instances in total.

As shown in Table~\ref{tab:model_performance_cnmo24}, CGPO achieves the largest average improvement across all evaluated models compared with other settings. In particular, for weaker models such as MetaMath-Mistral-7B and MetaMath-Llama-8B, CGPO improves performance from 0\% to a non-zero success rate. This shows that CGPO exhibits superior out-of-distribution generalization and can effectively enhance the performance of weaker models on challenging benchmarks, even without strong model guidance.

\subsection{Analysis of Scalability}

\begin{figure}[h] 
\centering 
\includegraphics[width=0.4\textwidth]{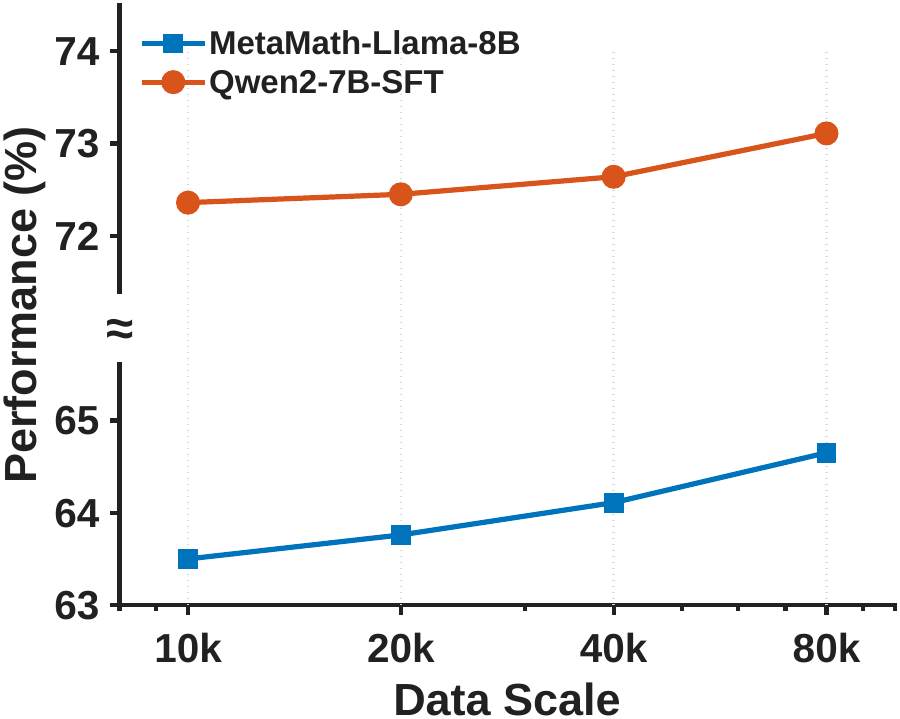} 
\caption{The relationship between the training dataset scale and the trained model performance.}
\label{fig:threshold1} 
\end{figure}

To explore model performance at different exploration degrees, we sampled 2, 4, and 8 times for each prompt on the Step-DPO-10k dataset using MetaMath-Llama-8B and Qwen2-7B-SFT, thereby constructing training datasets of 20k, 40k, and 80k. 

Figure~\ref{fig:threshold1} presents model performance on the GSM8K and MATH datasets. We find that increasing the exploration improves performance on both datasets, with a notable jump on GSM8K when the data volume increases from 40K to 80K. This suggests that increasing the training data enables models to explore more non-human-like reasoning paths, which further improves performance. These results indicate that CGPO exhibits effective scalability.

\subsection{Analysis of Reward Model}

To explore the impact of different reward models on CGPO, we adopt the process reward model proposed by Math-Shepherd \citep{wang-etal-2024-math} and Qwen-2.5-Math-PRM\citep{prmlessons} for comparison with ASPRM. 
To completely eliminate the influence of the reward model, we additionally compare against a variant that samples by directly selecting the tokens with the highest and lowest logits among the top-$k$ candidates.

\begin{table}[htbp]
\centering
\caption{Results of training datasets constructed by different reward models.}
\label{tab:model_performance_on_diff_reward_model}

\setlength{\tabcolsep}{5pt}
\renewcommand{\arraystretch}{1.0}
\normalsize

\begin{tabularx}{\columnwidth}{X c c}
\toprule
\textbf{Model} & \textbf{GSM8K} & \textbf{MATH} \\
\midrule

\textbf{MetaMath-Llama-8B} & 81.8 & 39.3 \\
\hspace{1em}+ Math-Shepherd      & 84.1 & 40.2 \\
\hspace{1em}+ Qwen-2.5-Math-PRM  & 84.1 & 40.7 \\
\hspace{1em}+ Logits              & 83.8 & 39.5 \\
\rowcolor{gray!15}
\hspace{1em}+ ASPRM              & 86.4 & 40.7 \\
\cmidrule(lr){1-3}

\textbf{Qwen2-7B-SFT} & 87.9 & 54.2 \\
\hspace{1em}+ Math-Shepherd      & 88.2 & 54.7 \\
\hspace{1em}+ Qwen-2.5-Math-PRM  & 88.3 & 55.0 \\
\hspace{1em}+ Logits              & 88.9 & 54.4 \\
\rowcolor{gray!15}
\hspace{1em}+ ASPRM            & 89.8 & 55.0 \\
\bottomrule
\end{tabularx}
\end{table}

The comparison results are presented in Table ~\ref{tab:model_performance_on_diff_reward_model}. When using Math-Shepherd or Qwen-2.5-Math-PRM as the reward model in CGPO, the trained models still outperform the base model. Without any reward model, training on reasoning paths sampled from the top-k tokens by logit scores still yields a modest gain over the base model. These results indicate that CGPO does not depend on a stronger reward model and that its improvements primarily come from exploring and optimizing non-human-like reasoning paths. The largest gains are observed with ASPRM, likely because its token-level discrimination supports more reliable selection of non-human-like paths.

\subsection{Analysis of Division Strategy}

We use confidence scores to reconstruct the initial reasoning step, as well as the chosen and rejected data in the original Step-DPO-10k dataset, in order to compare human-like and non-human-like step divisions. Specifically, we truncate the initial reasoning step in the Step-DPO-10k dataset at the lowest confidence point, as in the dataset construction of CGPO. For the chosen and rejected data, we apply a stopping threshold at the lowest 2\% confidence. 

\begin{table}[htbp]
\centering
\caption{Model performance with different annotation or split methods.}
\setlength{\tabcolsep}{5pt}
\renewcommand{\arraystretch}{1.0}

\begin{tabularx}{\columnwidth}{X c c}
\toprule
\textbf{Model} & \textbf{GSM8K} & \textbf{MATH} \\
\midrule

\textbf{MetaMath-Llama-8B} & 81.8 & 39.3 \\
\hspace{1em} + Step-DPO             & 84.8 & 40.2 \\
\hspace{1em} + Step-DPO-Confidence  & 84.4 & 40.2 \\
\rowcolor{gray!15}
\hspace{1em} + CGPO                & 86.4 & 40.7 \\
\cmidrule(lr){1-3}

\textbf{Qwen2-7B-SFT} & 87.9 & 54.2 \\
\hspace{1em} + Step-DPO             & 88.4 & 54.6 \\
\hspace{1em} + Step-DPO-Confidence  & 88.1 & 54.5 \\
\rowcolor{gray!15}
\hspace{1em} + CGPO               & 89.8 & 55.0 \\
\bottomrule
\end{tabularx}

\label{tab:step-dpo-confidence}
\end{table}

The final results are reported in Table~\ref{tab:step-dpo-confidence}. We find that even using $\pi_\theta$ to split the reasoning path on outputs generated by the strong model for training, the resulting performance still surpasses the baseline. However, it does not outperform training on the $\pi_\theta$ own generated and split data. This result suggests that the split of its own generated outputs by the weak model is more aligned with the model, thereby leading to improved training performance. 

\subsection{Analysis of Stopping Threshold}

\begin{figure}[h] 
\centering 
\includegraphics[width=0.4\textwidth]{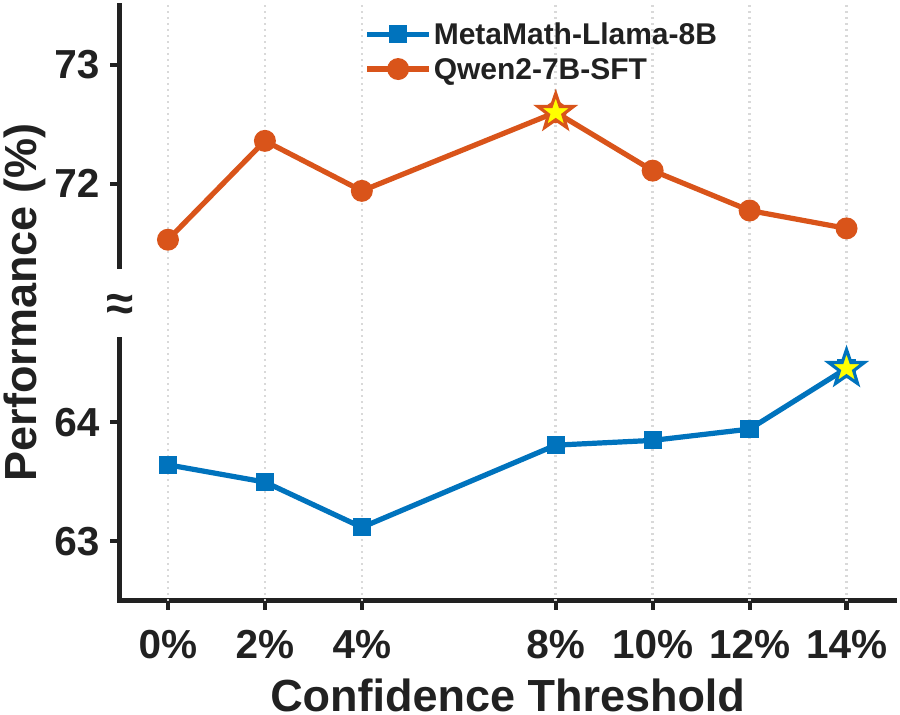} 
\caption{The relationship between the confidence threshold and the trained model performance.}
\label{fig:threshold} 
\end{figure}

We report the performance of the model under different confidence thresholds at which the sampling stops in Figure~\ref{fig:threshold}. For MetaMath-Llama-8B, performance keeps improving as the threshold increases. For Qwen2-7B-SFT, overall performance also shows an upward trend as the threshold rises, but the peak is achieved at a confidence threshold of 8\%. 

\begin{figure}[h]
\centering
\includegraphics[width=0.43\textwidth]{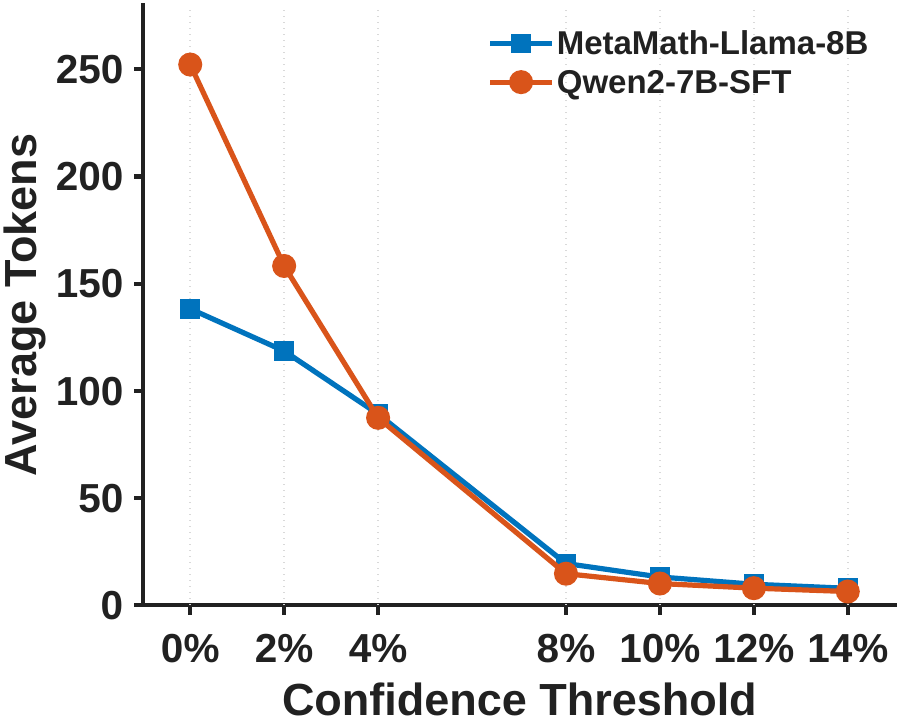}
\caption{The relationship between the confidence threshold and the average number of tokens per pair.}
\label{fig:ts-avgtoken}
\end{figure}

When we increase the confidence threshold, more samples will stop in the intermediate process rather than the generation of the EOS tag. This increases the proportion of non-human-like data, thereby enhancing the model's exploration and learning of non-human-like reasoning paths, which ultimately improves performance. However, increasing the confidence threshold also introduces shorter lengths of chosen and rejected data, as shown in Figure~\ref{fig:ts-avgtoken}. We argue that this is the reason for the performance decline of Qwen2-7B-SFT when the confidence threshold is greater. These results further demonstrate the necessity of optimizing the model by the intermediate reasoning path rather than by the entire process.

\section{Conclusion}

In this paper, we propose \MethodName, a DPO-based step-wise reasoning path preference optimization method. Our method emphasizes non-human-like reasoning paths in two key ways: 1) the training data does not rely on strong models; 2) the split for reasoning paths does not depend on predefined anchors. We show the effectiveness of \MethodName\ in both mathematics and code domains, achieving significant improvements and generalization ability over approaches that use data generated by strong models. Our analyses show that exploring non-human-like reasoning paths enhances model capabilities and is more effective than enforcing human-like reasoning paths. Separately, our experiments reveal that, within a certain range, intermediate reasoning path-focused optimization leads to greater performance gains than whole reasoning path optimization.

\clearpage

\section*{Impact Statements}

CGPO is an automated, scalable, and general method designed to optimize reasoning trajectories in LLMs through confidence-guided preference learning. By shifting away from labor-intensive human annotations, CGPO enables the autonomous generation and selection of high-quality reasoning paths, making it readily adaptable to a vast array of complex tasks. Our investigation into non-human-like, model-native reasoning paths offers a new perspective on optimization algorithms.

\bibliography{custom}
\bibliographystyle{icml2026}

\clearpage
\appendix

\clearpage          
\onecolumn         
\section{Appendix}
\label{sec:appendix}

\subsection{Example of Confusion vs. First Error}
\label{app:case_conf_error}

\begin{figure}[!h]
  \centering
  \includegraphics[width=1\textwidth]{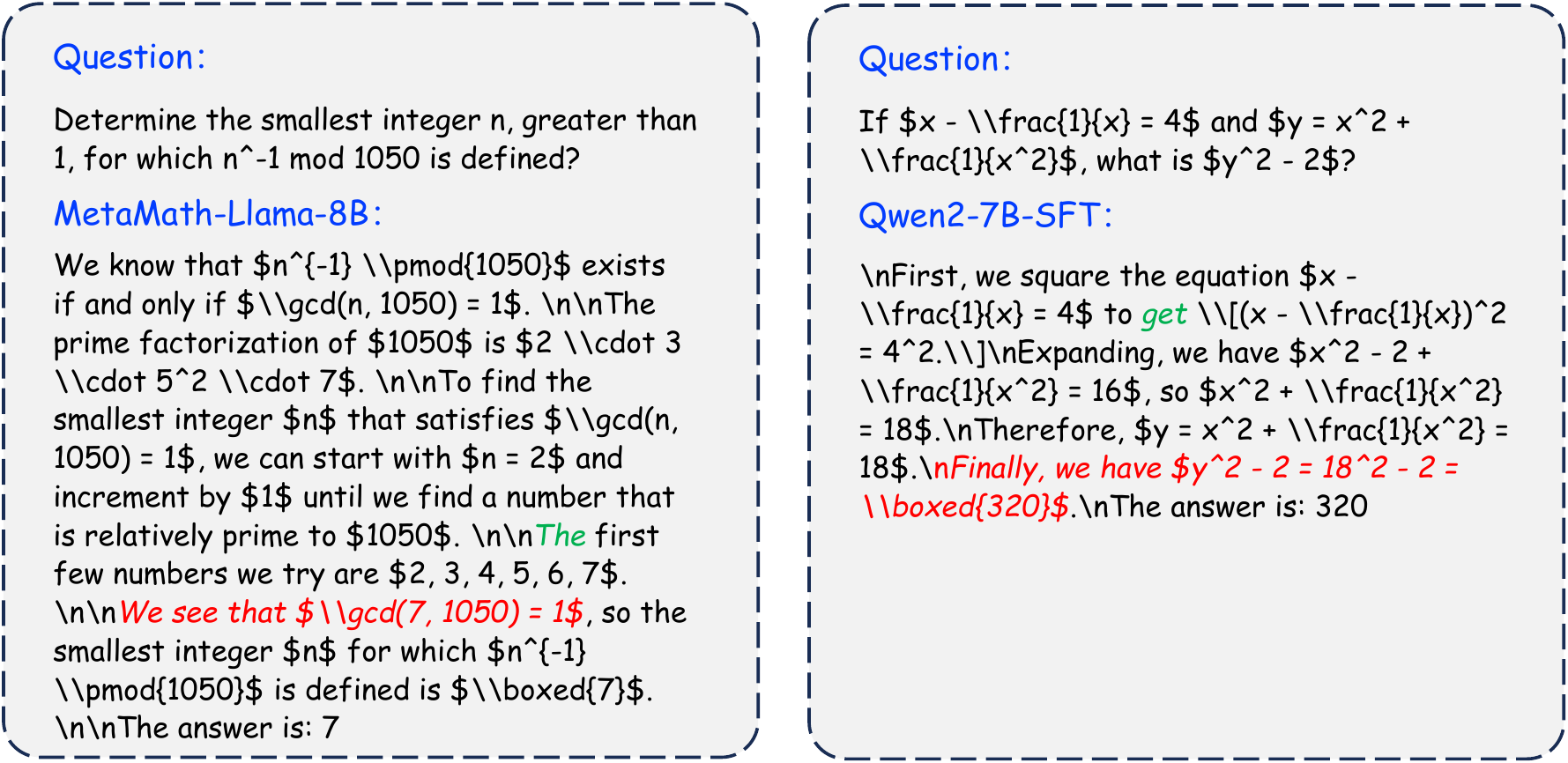}
  \caption{Examples where the lowest confidence appears before the first error.}
  \label{fig:case_conf_error}
\end{figure}

We present two examples where the lowest confidence appears prior to the first error in Figure~\ref{fig:case_conf_error}.

\subsection{Statistics of the Positional Relationship}
\label{app:static_error_confused_point}

\begin{table}[h]
\centering
\caption{Statistics on the relative positions of first error step and lowest confidence point, where "Before" and "After" denote the first error before or after the model confusion point}
  \renewcommand{\arraystretch}{1.25} 
  \vspace{6pt}
\begin{tabular}{lcc}
\toprule
\multirow{2}{*}{\textbf{Model}} & \multicolumn{2}{c}{\textbf{Position}} \\
\cline{2-3}
                       & Before & After \\
\midrule
MetaMath-Llama-8B         & 22     & 78    \\

Qwen2-7B-SFT                       & 28     & 72    \\
\bottomrule
\end{tabular}

\label{tab:model_position}
\end{table}

We randomly select 200 samples with incorrect results, half of which are from the MetaMath-Llama-8B generated CGPO training dataset and the other from the Qwen2-7B-SFT dataset. To identify the first erroneous step, we use GPT5-thinking with the prompt-

\textit{"Analyze the following text and identify the first location where an error occurs, reporting the exact substring/token, its position (line and character index), why it is an error, and why this is the first error (i.e., why all preceding text is error-free); if no error exists, reply `No error found'; Text:" [Q-A pair]}

-to find the first error, and manually determine the positional relationship between the first error and the model's lowest confidence point. We report the analysis results in Table~\ref {tab:model_position}.

\subsection{Training Details}
\label{sec:training_details}

The key hyperparameters we used in training the model are presented in Table~\ref{tab:training_config}, including the learning rate, beta, batch size, and number of epochs. For all models with fewer than 32B parameters, we set $\beta$ to 0.1, while for 32B model, we set $\beta$ to 0.3.

\begin{table}[h]
\centering
\caption{Training hyperparameters for all models.}
\label{tab:training_config}
\begin{tabular}{lcccc}
\toprule
Model & $\beta$ & LR & Batch Size & Epochs \\
\midrule
MetaMath-Mistral-7B & 0.1 & $5\times10^{-7}$ & 128 & 4 \\
MetaMath-Llama-8B & 0.1 & $5\times10^{-7}$ & 128 & 4 \\
Qwen2-7B-SFT & 0.1 & $5\times10^{-7}$ & 128 & 4 \\
DeepSeekMath-Base-7B-SFT & 0.1 & $5\times10^{-7}$ & 128 & 4 \\
Qwen1.5-32B-SFT & 0.3 & $5\times10^{-7}$ & 128 & 4 \\
DeepSeek-R1-Distill-Qwen-1.5B & 0.1 & $5\times10^{-7}$ & 128 & 4 \\
DeepSeek-R1-Distill-Qwen-7B & 0.1 & $5\times10^{-7}$ & 128 & 4 \\
Deepseek-Coder-7B-Instruct-v1.5 & 0.1 & $5\times10^{-7}$ & 128 & 4 \\
\bottomrule
\end{tabular}
\end{table}

\subsection{CGPO performance on different sampling temperatures}

\begin{figure}[h] 
\centering 
\includegraphics[width=0.45\textwidth]{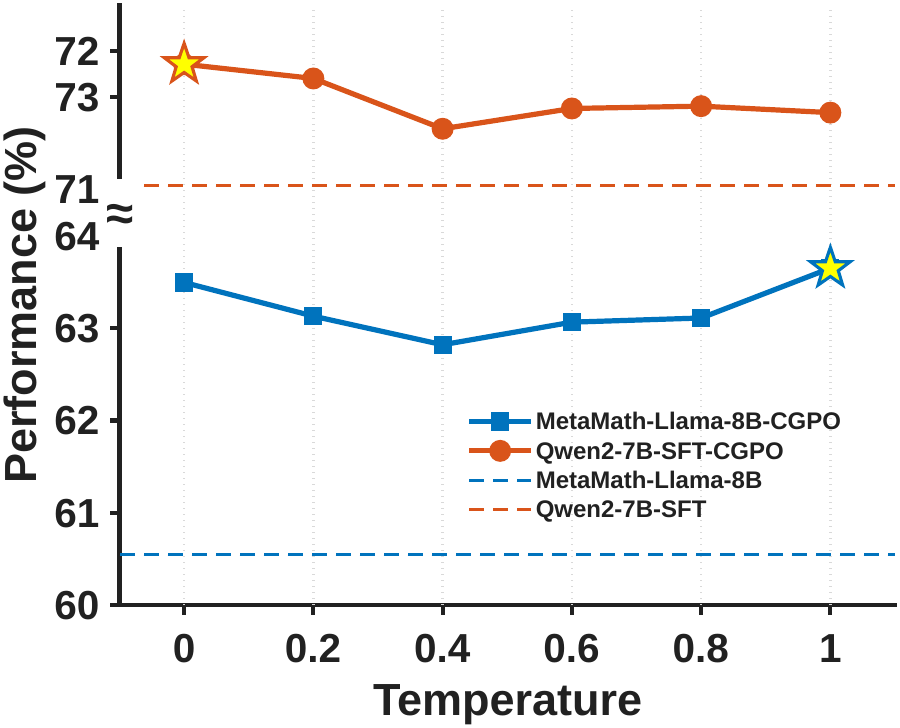} 
\caption{The relationship between the sampling temperatures and the trained model performance.}
\label{fig:temperature} 
\end{figure}

We experimented with using different temperatures in CGPO to sample non-human-like data. The figure~\ref{fig:temperature} show that, across all evaluated temperatures, CGPO consistently outperforms the base model, demonstrating its robustness to the sampling temperatures.

\subsection{CGPO Training Data Analysis}
\label{sec:data_analysis}

\begin{table}[htbp]
\centering
\small
\caption{Samples and Proportion of decision tokens for CGPO training dataset.}
\label{tab:lowest_confidence_token_categories}
\renewcommand{\arraystretch}{1.3} 
\begin{tabular}{l c c l}
\toprule
\textbf{Category} & \textbf{Lowest-conf. (\%)} & \textbf{All Tokens (\%)} & \textbf{Sample Snippet} \\
\midrule
\multicolumn{4}{l}{\textbf{\textit{Math Reasoning}} }\\
\hspace{1em} Number & 17.99 & 17.45 & sum is an integer, the answer is \textcolor{red}{2000}. \\
\hspace{1em} Proper Noun & 8.30 & 4.44 & 7000 miles: Bangkok\textcolor{red}{-Honolulu} \\
\hspace{1em} Symbol & 9.45 & 8.58 & 90 + 90 + ... + 71 \textcolor{red}{=} 420 degrees. \\
\hline
\addlinespace
\multicolumn{4}{l}{\textit{\textbf{Logical Relation}}} \\
\hspace{1em} Conjunction & 9.15 & 3.97 & We \textcolor{red}{then} add and subtract \$25\$ \\
\hspace{1em} Verb & 9.10 & 7.74 & we \textcolor{red}{calculate} the sum of 46 and 37 \\
\addlinespace
\hline
\multicolumn{4}{l}{\textit{\textbf{Supportive Elements}}} \\
\hspace{1em} Particle & 4.29 & 1.51 & positive integers that do \textcolor{red}{not} have \\
\hspace{1em} Auxiliary Verb & 0.50 & 5.97 & we \textcolor{red}{have} $y^2 + 12y = 45$ \\
\midrule
\textbf{Others} & 41.22 & 50.34 & -- \\
\bottomrule
\end{tabular}
\end{table}

We categorize 21,374 lowest-confidence tokens using a rule-based lexical taxonomy into three categories: Math Reasoning, Logical Relation, and Supportive Elements. 
Table~\ref{tab:lowest_confidence_token_categories} presents the distribution of these decision-related token categories in the CGPO training dataset compared to their overall frequency in all tokens. We observe that tokens associated with reasoning and computation are significantly enriched at the model's lowest-confidence points. Specifically, numeric tokens, computation symbols, and computation-related verbs and prepositions occur more frequently among decision tokens than in the overall token distribution. This enrichment indicates that the lowest-confidence points selected by CGPO are strongly correlated with reasoning-related operations.

\end{document}